\documentclass{bmvc2k}

\title{Identifying Individual Dogs \\ in Social Media Images}

\addauthor{Djordje Batic}{djordjebatic@gmail.com}{1}
\addauthor{Dubravko Culibrk}{http://www.dubravkoculibrk.org}{1}
\addinstitution{
 Faculty of Technical Sciences\\
 University of Novi Sad\\
 Novi Sad, Serbia
}

\runninghead{Batic, Culibrk}{Identifying Individual Dogs in Social Media}

\def\etal{\emph{et al}\bmvaOneDot}

\begin{document}

\maketitle

\begin{abstract}
We present the results of an initial study focused on developing a visual AI solution able to recognize individual dogs in unconstrained (wild) images occurring on social media. 

The work described here is part of joint project done with Pet2Net, a social network focused on pets and their owners. In order to detect and recognize individual dogs we combine transfer learning and object detection approaches on Inception v3 and SSD Inception v2 architectures respectively and evaluate the proposed pipeline using a new data set containing real data that the users uploaded to Pet2Net platform. We show that it can achieve 94.59\% accuracy in identifying individual dogs. Our approach has been designed with simplicity in mind and the goal of easy deployment on all the images uploaded to Pet2Net platform.

A purely visual approach to identifying dogs in images, will enhance Pet2Net features aimed at finding lost dogs, as well as form the basis of future work focused on identifying social relationships between dogs, which cannot be inferred from other data collected by the platform.

%A purely visual approach to identifying dogs in images, will enhance Pet2Net features aimed at finding lost dogs, as well as form the basis of future work focused on identifying social relationships between dogs, which cannot be inferred from other data collected by the platform.

\end{abstract}

%-------------------------------------------------------------------------
\section{Introduction}
\label{sec:intro}
The American Pet Products Association (APPA) estimates that 78 million dogs are owned in the US. A recent study \cite{weiss2012frequency} showed that 15\%
 of dogs get lost, with 7\%, i.e. 819,000 of that never recovered. Development of a reliable system for visual dog identification, combined with social media could help recover some of these dogs, as well as pave the way towards deeper understanding of social networks of both pets and their owners. 
%72.6 billion dollars has been spent on pets and related products and services 8 in US alone. The association 

Pet2Net is a social platform focused on pets and their owners looking to develop and leverage visual AI tools to enhance the online experience of both its human users and their animals. This study represents an initial foray into the analysis of images uploaded to the platform aimed at identifying individual dogs. In the first instance we hope to develop an automatic tool as an aid to recognizing lost pets that would complement the functionality already offered by the platform. As the next step the tool will be routinely deployed to infer social dynamics of dogs present in the images uploaded to the platform, which cannot be derived from other data provided by the users.  

There has been significant effort in the computer vision community focused on identifying dog breeds, spurred by the creation of the Stanford dogs data set as a benchmark for fine-grained visual object classification \cite{khosla2011novel} and the Kaggle dog breed identification challenge \cite{Kaggle}. Surprisingly, we have been able to identify just two recent studies focusing of recognizing specific dogs \cite{moreira2017my}\cite{tu2018transfer}. Both focus on the animals' faces as a discerning biometric and are trained (or rather fine tuned) using a relatively small set (Flickr-dog) collected by Moreira \etal \cite{moreira2017my}. The images were pre-processed to crop the dog faces, rotate them to align the eyes horizontally, and resize them to $250 \times 250$ pixels. Thus reducing the problem to image classification Moreira and \etal explored a few approaches, achieving the best result using OverFeat deep neural network \cite{sermanet2013overfeat} to extract features, which were subsequently fed into an SVM classifier. The classifier achieves 66.9\% accuracy on this data set. Moreira \etal conducted additional experiments using a mongrel data set containing 18 dogs, on which the same approach achieved significantly higher results (89.4\%), but this data set is not openly available.

Tu \etal \cite{tu2018transfer} proposed a dog identification approach that relies purely on deep learning. They used both the Stanford dogs\cite{khosla2011novel}, as well as the Columbia dogs with parts data set\cite{liu2012dog}. The data set from Columbia contains facial markers which were used by Tu \etal to generate bounding boxes of the faces train a Faster-RCNN-based \cite{ren2015faster} dog face detector. The detector was then used to detect the faces within the bounding boxes of dogs already present in the Stanford data set, creating a combined data set containing 25,798 images of dog faces, covering 198 breeds, but with labels indicating only the breed, rather than individual dogs. This combined data set was used to fine tune an architecture based on GoogLeNet \cite{szegedy2015going} (BreedNet) to recognize breeds from the faces of dogs. Finally, the images of dogs in the Flickr-dog were augmented by flipping them along the vertical axis and used to finetune BreedNet to recognize individual dogs, achieving an accuracy of 83.94\%. Using an entirely deep-learning-based approach lead to significant improvement of accuracy on the small Flickr-dog data set, even if the BreedNet could not achieve state-of-the-art accuracy in terms of breed classification.

When it comes to how good humans are at recognizing individual dogs, a study conducted by Diamond and Carey \cite{diamond1986faces} found that humans were only 76\% accurate when identifying dogs in optimal conditions, but when viewing images of similar dogs of three different species in show stances, not only faces. When the subjects were experts (breeders and judges from the American Kennel Club/New York) the accuracy rose to 81\%. Thus, recognizing a large number of dogs is far from trivial, even for human experts and they typically do not rely on faces alone for dog identification.

In the study presented here, we developed an automatic dog recognition module intended for the Pet2Net platform and trained and evaluated it using a data set of real images collected from the Platform. The work presented here is more closely related to that of Tu \etal \cite{tu2018transfer} than to that of Moreira \etal \cite{moreira2017my}, as we also used a deep neural network and transfer learning to create our classifier. However, contrary to both approaches, we took a more straight forward approach and base our identification on the whatever parts of the dog are visible, instead of focusing on faces alone. In addition,  since our data is "wild", incorporating not only dogs, but humans and other animals as well, and we have only image-level annotations indicating that a specific dog is present in the image, we first needed to implement a module to extract the portions of the images containing dogs and then apply our dog identification module. Also, we did not make any specific attempts to pre-process and normalize the data either manually or automatically based on face features previously annotated by humans, as was done by Moreira \etal and Tu \etal, respectively. This makes the proposed approach much more suitable for applications on large data sets of images, such as those that can reasonably be encountered in social networks like the Pet2Net.   

\section{Proposed approach and methodology}
\begin{figure}[h]
\centering
\includegraphics[width=0.8\textwidth, height=4.5cm]{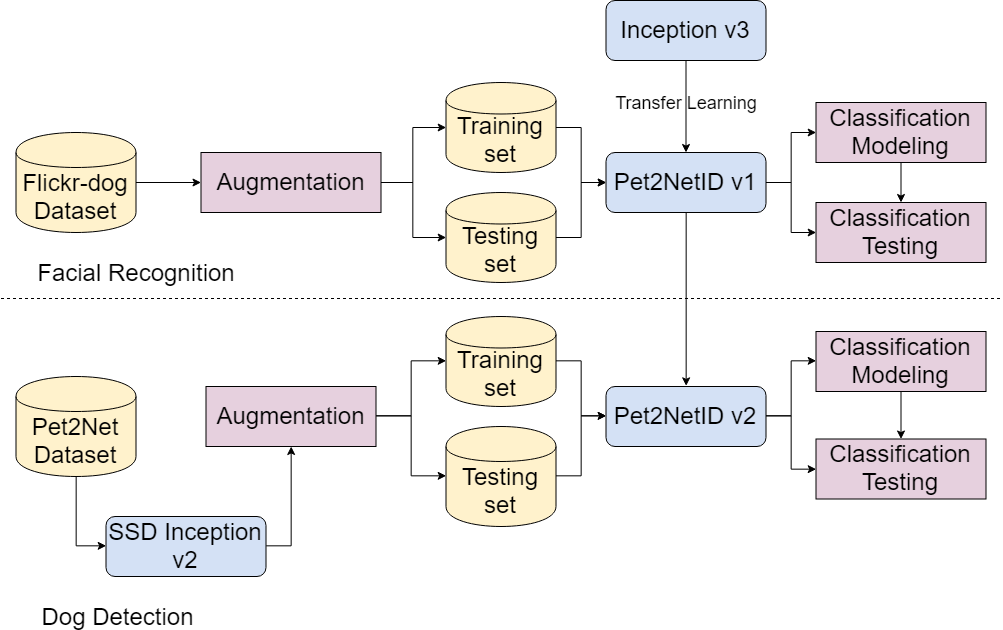}
\caption{Proposed pipeline. The inputs are images in our data sets and the output is the predicted identity of dogs.}
\label{fig:pipeline}
\end{figure}

The pipeline of our approach is shown in Figure \ref{fig:pipeline}. To create the initial version of our Pet2Net Identification (Pet2NetID) network, we first designed a model based on the Inception v3 architecture \cite{szegedy2016rethinking}, trained on IMAGENET data set \cite{imagenet_cvpr09} and then finetuned and evaluated it using the Flickr-dog data set. This provided our Pet2NetID v1 model. Our architecture was created by extending the Inception v3 architecture with a global average pooling layer and a fully connected layer containing 1024 neurons. Dropout (0.5 likelihood) was used during training to limit overfitting. Since the data set is small, we used the image augmentation to enlarge it. Images were flipped, shifted, sheared and zoomed in to create the training data set which was 16 times larger than the original.   

We then used transfer learning the adapt the model to Pet2Net data. Before feeding the data into the Pet2NetID net, we needed to detect the dogs in the images, so we used Single Shot Multibox Detection (SSD) \cite{liu2012dog} based on the Inception v2 architecture to do that. Once the bounding boxes for dogs were available, we extracted 3 square overlapping windows along the longer dimension of the bounding box, labelled them with the individual dog ID and used this data to finetune our Pet2NetID v1 model and create our final Pet2NetID (v2) network. The same data augmentation was used as in the Flickr-dog case.

At inference time an image is fed to the SSD network first to extract the bounding boxes, split into three square windows and each of them fed to the Pet2NetID. We then use majority voting to determine the final label. In case there in no majority label, the label with the strongest activation is selected.   

\subsection{Data}
We employed two data sets, Pet2Net and Flick-dog in our experiments.  

The Flickr-dog data set contains two breeds of dogs: pug and husky. There are 21 individual dogs with at least 5 images for each dog, totaling 374 photos of
42 subjects. The dogs' face images are cropped, rotated to align the eyes horizontally, and resized to 250x250 pixels.

\begin{figure}[h]
\begin{tabular}{cc}
\bmvaHangBox{\fbox{\includegraphics[height=4cm]{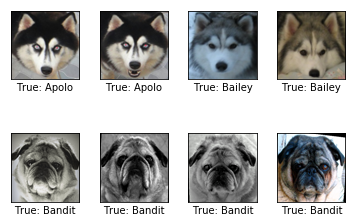}}}&
\bmvaHangBox{\fbox{\includegraphics[height=4cm]{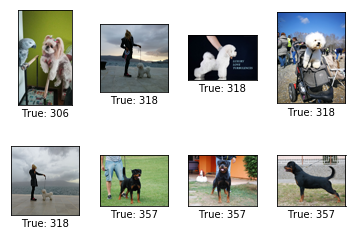}}}\\
(a)&(b)
\end{tabular}
\caption{Samples of the dog images and their true labels in two data sets: Flickr-dog (a) and Pet2Net (b) contain various individual dogs of two breeds (pugs and huskies) and multiple different breeds, respectively.}
\label{fig:data}
\end{figure}

The Pet2Net data set is new data set collected from the Pet2Net social network. There are 16 individual dogs with at least 5 images each, totaling 213 images. The data includes 10 different breeds, including some that do not exist in the IMAGENET and, therefore, Stanford Dogs data set, such as Dogo Argentino. Many images are either showing the full figure of the dog, several dogs, or contain not just dogs, but humans and other animals as well. Sample images from the two data sets are shown in Figure \ref{fig:data}.

\section{Results}
All our models were implemented and our experiments conducted using TensorFlow \cite{tensorflow2015-whitepaper}. We used built-in functionality for both data augmentation and SSD-based-on-Inception-v2 object detection. 
%The object detection module extracted the bounding boxes of objects recognized as dogs with over 85\% certainty. 

We first trained our Pet2NetID network using the Flickr-dog data set and achieved 10-fold cross validation accuracy of 77.19\%, which is better than that reported by Moreira \etal (66.9\%). Tu \etal managed to achieve 83.94\%, but only using an additional Culumbia Dogs with Parts data set \cite{liu2012dog} and a significantly more complex pipeline.  

We then used SSD to detect dogs in the Pet2Net images. The crops corresponding to detected dogs were then used to extract 3 square overlapping windows per crop, which formed the basis of the data set used to finetune our Pet2NetID network. The same augmentation procedure was used as in the Flickr-dog data set and our network achieved 94.59\% accuracy at identifying individual dogs in our data set. 
\section{Conclusion}
We presented an initial study focused on developing a visual AI solution for dog identification based on images shared by the users of the Pet2Net social network. The developed approach is able to identify individual dogs in unconstrained images, commonly occurring on Pet2Net and other social media and does not require any data other than that available on the platform for training, making it suitable for deployment at scale. In the experiments conducted within this study, the proposed approach achieved an accuracy of 94.59\%, surpassing human-level performance. The module will routinely be deployed on all images uploaded on the platform in the future, enhancing the pet finding features of Pet2Net and paving the way for complex dog relationship analyses.     

\bibliography{egbib}
\end{document}